# Prompt engineering paradigms for medical applications: scoping review and recommendations for better practices

Jamil Zaghir‡*[12], Marco Naguib‡[3], Mina Bjelogrlic[12], Aurélie Névéol[3], Xavier Tannier[4], Christian Lovis[12]

‡Equal contribution

[1]Division of Medical Information Sciences, Geneva University Hospitals, Geneva, Switzerland

[2]Department of Radiology and Medical Informatics, University of Geneva, Geneva, Switzerland

[3]Université Paris-Saclay, CNRS, Laboratoire Interdisciplinaire des Sciences du Numérique, 91400, Orsay, France

[4]Sorbonne Université, INSERM, Université Sorbonne Paris-Nord, Laboratoire d'Informatique Médicale et d'Ingénierie des Connaissances en eSanté, LIMICS, Paris, France

## Abstract

Prompt engineering is crucial for harnessing the potential of large language models (LLMs), especially in the medical domain where specialized terminology and phrasing is used. However, the efficacy of prompt engineering in the medical domain remains to be explored. In this work, 114 recent studies (2022-2024) applying prompt engineering in medicine, covering prompt learning (PL), prompt tuning (PT), and prompt design (PD) are reviewed. PD is the most prevalent (78 articles). In 12 papers, PD, PL, and PT terms were used interchangeably. ChatGPT is the most used LLM, with seven papers using it for processing sensitive clinical data. Chain-of-Thought emerges as the most common prompt engineering technique. While PL and PT articles typically provide a baseline for evaluating prompt-based approaches, 64% of PD studies lack non-prompt-related baselines. We provide tables and figures summarizing existing work and reporting recommendations to guide future research contributions.

## Introduction

In recent years, the development of Large Language Models (LLMs) such as GPT-3 have disrupted the field of Natural Language Processing (NLP). LLMs have demonstrated capabilities in processing and generating human-like text, with applications ranging from text generation and translation to question answering and summarization [1]. However, harnessing the full potential of LLMs requires careful consideration of how input prompts are formulated and optimized [2].

Input prompts denote a set of instructions provided to the LLM to execute a task. Prompt Engineering, a term coined to describe the strategic design and optimization of prompts for LLMs, has emerged as a crucial aspect of leveraging these models. By crafting prompts that effectively convey tasks or queries, researchers and practitioners can guide LLMs to improve accuracy and pertinence of responses. The literature defines prompt engineering in various ways: it can be regarded as a prompt structuring process that enhances the efficiency of an LLM to achieve a specific objective [3], or as the mechanism through which LLMs are programmed by prompts [4]. Prompt engineering encompasses a plethora of techniques, often separated into distinct categories such as output customization and prompt improvement [4]. Existing prompt paradigms are presented in more detail in the Methods section.

*Corresponding author - contact: jamil.zaghir@unige.ch

In the realm of medical NLP, significant advancements have been made, such as the release of LLMs specialized in medical language and the availability of public medical datasets, including in languages other than English [5]. The unique intricacies of medical language, characterized by its terminological precision, context sensitivity, and domain-specific nuances, demand a dedicated focus, and exploration of NLP in healthcare research. Despite these imperatives, to our knowledge, there is currently no systematic review analyzing prompt engineering applied to the medical domain.

The aim of this scoping review is to shed light on prompt engineering as it is developed and used in the medical field, by systematically analyzing the literature in the field. Specifically, we examine the definitions, methodologies, techniques, and outcomes of prompt engineering across various NLP tasks. Methodological strengths, weaknesses, and limitations of the current wave of experimentation is discussed. Finally, we provide guidelines for comprehensive reporting of prompt engineering-related studies to improve clarity and facilitate further research in the field. We aspire to furnish insights that will inform both researchers and users about the pivotal role of prompt engineering in optimizing the efficacy of LLMs. By gaining a thorough understanding of the current landscape of prompt engineering research, we can pinpoint areas warranting further investigation and development, thereby propelling the field of medical NLP forward.

In this review, we use terminology to denote emerging technical concepts that lack consensus definitions. We propose the following definitions, based on previous use in the literature:

- **LLM**: Object that models language and can be used to generate text, by receiving large-scale language modeling pretraining (Luccioni and Rogers [6] defines an arbitrary threshold at 1B tokens of training data). An LLM can be adapted to downstream tasks through transfer learning approaches such as fine-tuning, or prompt-based techniques. Following Thirunavukarasu et al.'s study of models for the medical field [7], we include BERT-based and GPT-based models in this definition, although Zhao et al. [8] place BERT models in a separate category.
- **Fine-tuning**: Approach in which the weights of the pre-trained LLM are retrained on new samples. The additional data can be labeled and designed to adapt the LLM to a new downstream task.
- **Prompt Design (PD)** [1,2]: Manually building a prompt (named manual prompt, or hard prompt) tailored to guide the LLM towards resolving the task by simply predicting the most probable continuity of the prompt. The prompt is usually a set of task-specific instructions, occasionally featuring a few demonstrations of the task.
- **Prompt Learning (PL)** [3]: Manually building a prompt and passing it to an LLM, trained via the Masked Language Modeling (MLM) objective, to predict masked tokens. The prompt often features masked tokens over which the LLM makes predictions. Those are then projected as predictions for a new downstream task. This approach is also referred to as prompt-based learning.
- **Prompt Tuning (PT)** [9]: Refers to the LLM prompting where part or all the prompt is a trainable vectorial representation (as known as continuous prompt or soft prompt) that is optimized with respect to the annotated instances.

Figure 1 illustrates the four approaches described above.

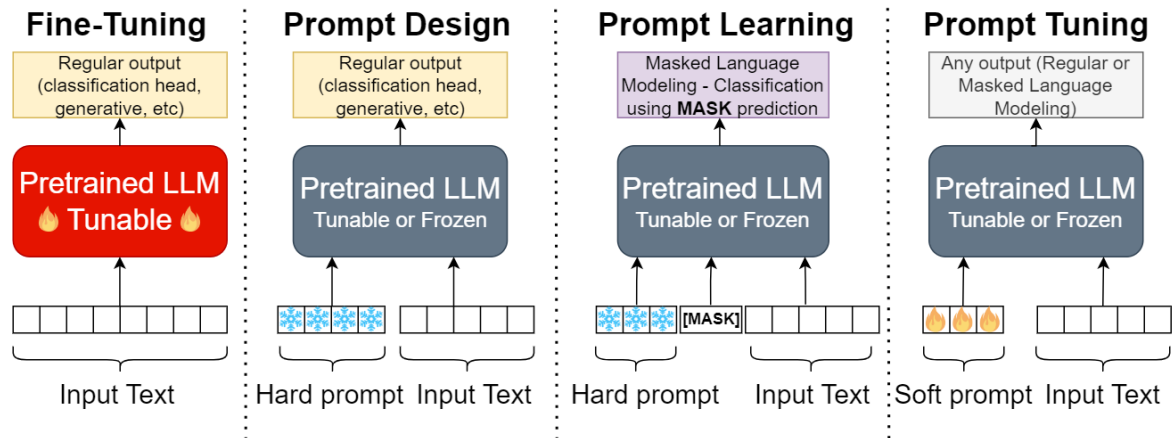

Figure 1: Illustration of traditional fine-tuning and the three prompt-based paradigms *(fire=trainable, flake=frozen)*

## Results

### Screening results

The systematic search across sources yielded 398 articles. Following the removal of duplicates, 251 articles underwent screening based on title, abstract and keywords, leading to the exclusion of 94 studies. During this first screening step, 33 conflicts were identified and resolved among the annotators, resulting in an inter-annotator agreement of 86.8%. Subsequently, 157 studies remained, and full-text copies were retrieved and thoroughly screened. This process culminated in the inclusion of a total of 114 articles in this scoping review. The detailed process of study selection is shown in Figure 2. Among the selected papers, 13 are from clinical venues, 33 are from medical informatics sources, 31 are from computer science publications and four are from other sources. Notably, 33 of them are preprints.

### Prompt paradigms and medical subfields

Table 1 depicts the number of articles identified within each prompt paradigm along with their associated medical subfields. Some articles may simultaneously involve several (up to two in this review) prompt paradigms. Notably, PD emerged as the predominant category, with a total of 78 articles. These articles spanned across various medical fields, with a greater emphasis on clinical (including specialties) rather than biomedical disciplines. The screening yields 29 PL articles and 19 PT articles, with both paradigms maintaining a balanced distribution between biomedical and clinical domains. However, it is noteworthy that unlike PL and PT, PD encompassed a much broader spectrum of clinical specialties, with a particular interest in Psychiatry.

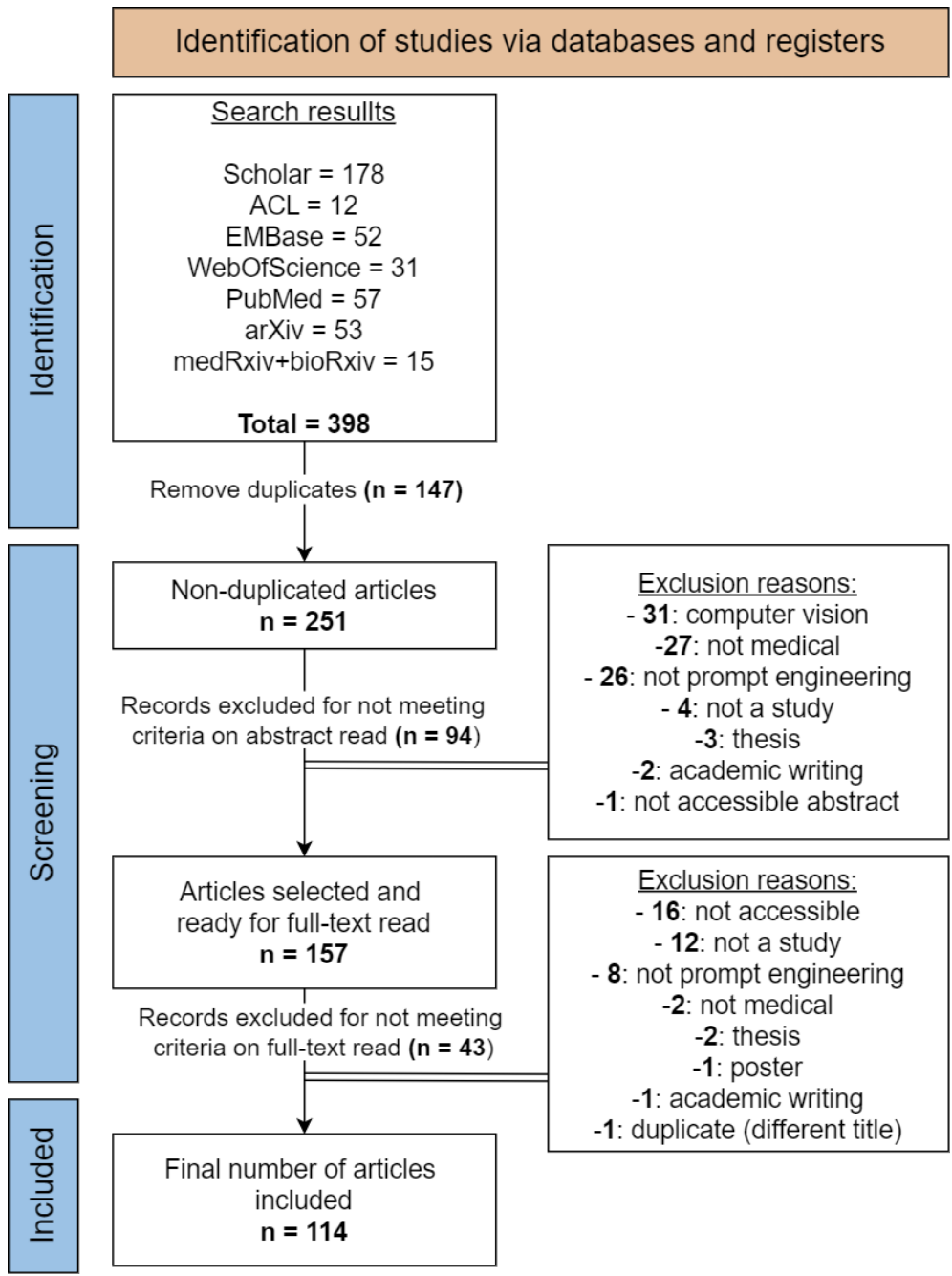

Figure 2: PRISMA flow diagram for the review process.

Table 1: Article distribution by prompt category and medical subfield, with corresponding references

| Prompt Paradigm | Domain of the topic | References |
|---|---|---|
| Prompt Design (78) | Biomedical (17) | [10–26] |
| | Medical Licensing Exam (12) | [27–38] |
| | Clinical (General) (15) | [39–53] |
| | Psychiatry (10) | [28,54–62] |
| | Oncology (5) | [63–67] |
| | Cardiology (4) | [68–71] |
| | Ophthalmology (3) | [72–74] |
| | Neurology (3) | [69,75,76] |
| | Orthopedics (2) | [77,78] |
| | Clinical Trials (2) | [79,80] |
| | Intensive care (2) | [69,81] |
| | Geriatrics (2) | [75,76] |
| | Radiology (2) | [31,82] |
| | Nuclear medicine (1) | [29] |
| | Hepatology (1) | [83] |
| | Endocrinology (1) | [84] |
| | Plastic surgery (1) | [85] |
| | Gastroenterology (1) | [32] |
| | Genetics (1) | [86] |
| | Nursing (1) | [87] |
| Prompt Learning (29) | Biomedical (13) | [88–100] |
| | Clinical (General) (15) | [41,47,101–113] |
| | Psychiatry (1) | [114] |

| Prompt Tuning (19) | Biomedical (9) | [16,20,26,90,91,95,98,115,116] |
| --- | --- | --- |
| | Clinical (General) (6) | [101,105,110,117–119] |
| | Oncology (2) | [120,121] |
| | Psychiatry (1) | [122] |
| | Medical Insurance (1) | [123] |

## Terminology usage

In our review, the consistency of terminology usage around prompt engineering was investigated, particularly concerning its three paradigms: PD, PL and PT. Across the articles, we meticulously tracked instances where the terminology was applied differently to the definitions used in the literature and described in the introduction. Notably, PL was used to refer to PD four times [12,13,67,86] and PT once [119], while PT was used five times to describe PL [88,96,97,99,114] and twice for PD [23,43]. Terminology inconsistencies were identified in only 12 studies. Consequently, while there remains some degree of inconsistency, a significant majority of 102 articles adhered to the definitions identified as commonly used terminology.

## Language of study

Considering the latest developments in NLP research encompassing languages beyond English [124], reporting the language of study is crucial. Several articles do not explicitly state the language of study. In some cases, the language can be inferred from prompt illustrations or examples. In the least informative cases, only the dataset of the study is disclosed, indirectly hinting at the language.

Figure 3 illustrates the language distribution among the selected articles, noting whether languages are explicitly mentioned, implicitly inferred from prompt illustrations, or simply not stated but implied from the employed dataset. The language used in two articles [60,68] remains unknown.

Notably, English dominates with 84.2% of the selected articles, followed by Chinese at 15.8%. Then, the other languages are relatively rare, often appearing in studies featuring multiple languages. It is worth mentioning that languages besides English are usually explicitly stated, with the exception of an article studying Korean [63]. In total, the language had to be inferred from prompt figures and examples in 48 papers, all in English.

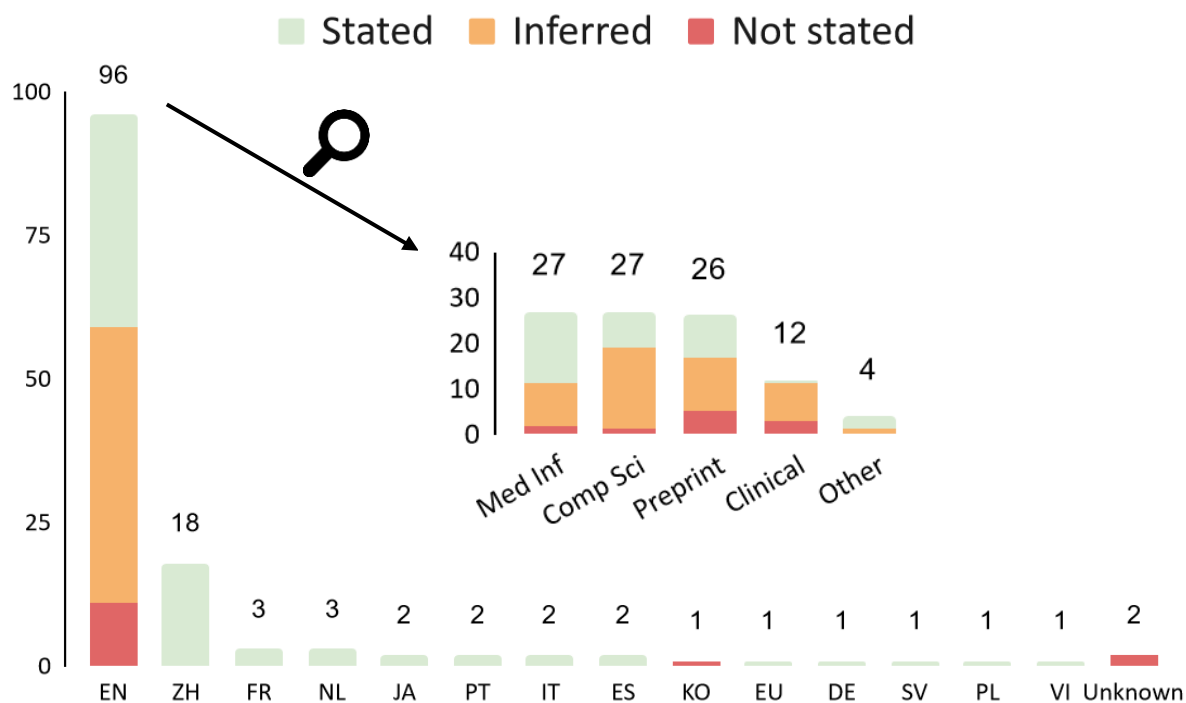

Figure 3: Frequency distribution of articles across various languages (as ISO 639 language codes). *Stated* = stated in the article. *Inferred* = inferred from prompt figures and examples. *Non stated* = inferred from the dataset. The smaller figure depicts the frequency distribution across venues for articles studying English.

## Choice of Large Language Models

Given the diverse array of LLMs available, spanning general or medical, open-source or proprietary, and monolingual or multilingual models, alongside various architectural configurations (encoder, decoder, or both), our study investigates LLM selection across prompt paradigms.

Figure 4 outlines prevalent LLMs categorized by prompt paradigms, though it is not exhaustive and only includes commonly encountered architectures. For example, while encoder-decoder models are absent in PT in Figure 4, there are a few instances where they are utilized [95,110].

ChatGPT's popularity in PD is unsurprising, given its accessibility. Models from Google, PaLM and Bard, all falling under closed models, are also prominent. Among open-source instruct-based LLMs, fewer are employed, notably those based on LLaMA-2 with 7 occurrences.

In PL, encoder models, those following the BERT architecture, dominate, covering both general and specialized variants. While Figure 4 does not display it, there are occasional uses of decoder models like GPT-2 in PL-based tasks [103,105]. PT involves all model types, with a preference towards encoders. Further details on the models used are available in Supplementary Material.

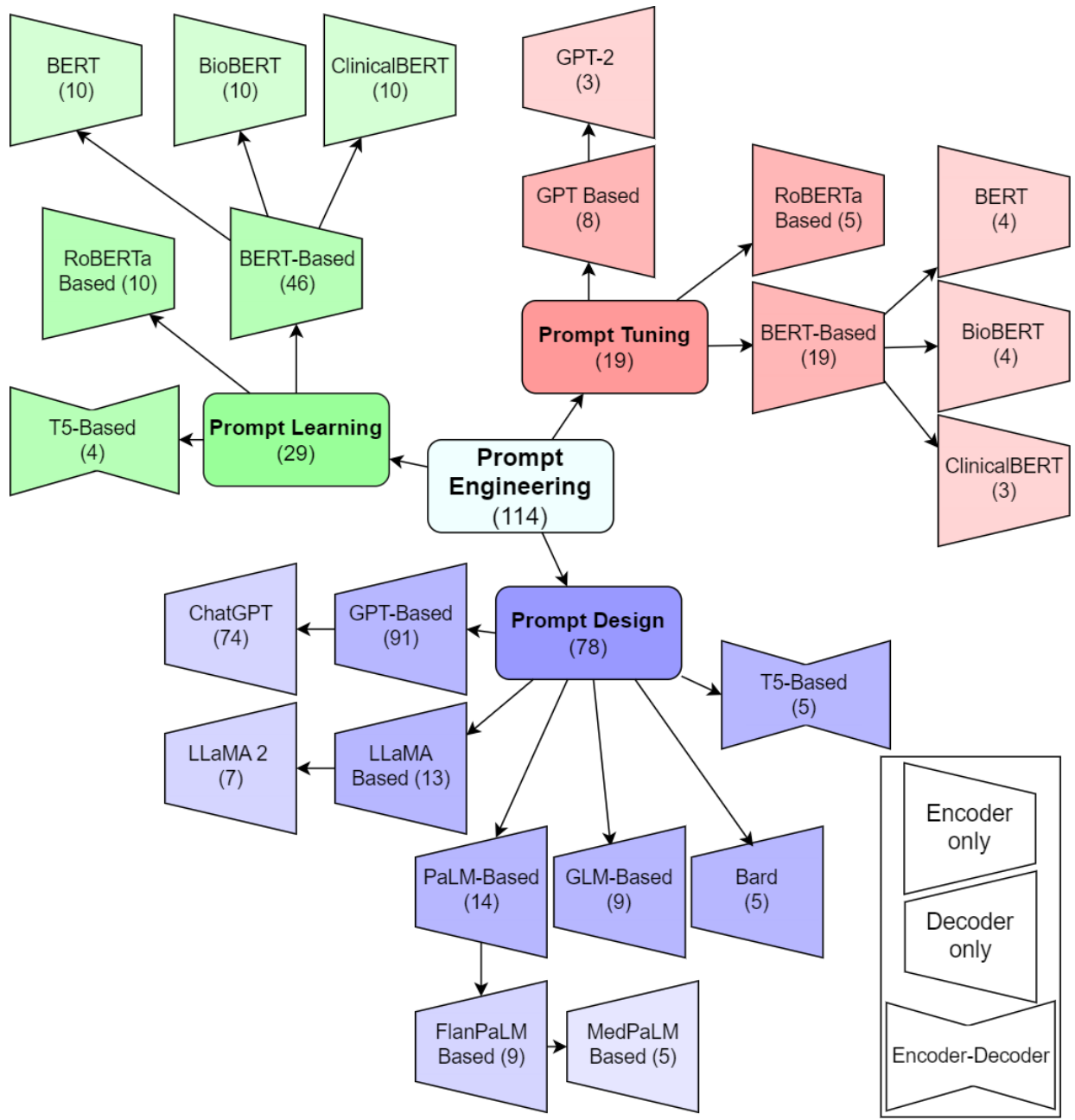

Figure 4: Involved LLMs in the prompt engineering studies, covering all prompt paradigms. The number of studies that fit in a node is shown in parenthesis.

### Topic domain and NLP tasks trends

Figure 5 illustrates the target tasks utilized in the PL and PT articles. PL-focused papers predominantly address classification-based tasks such as text classification, named entity recognition, and relation extraction, with text classification being particularly prominent. This aligns with the nature of PL, which centers around an MLM objective. Among other tasks, a study based on text generation [111] makes use of PL to predict masked tokens from partial patient records, aiming to generate synthetic electronic health records. Conversely, PT articles tend to exhibit a slightly broader range of tasks.

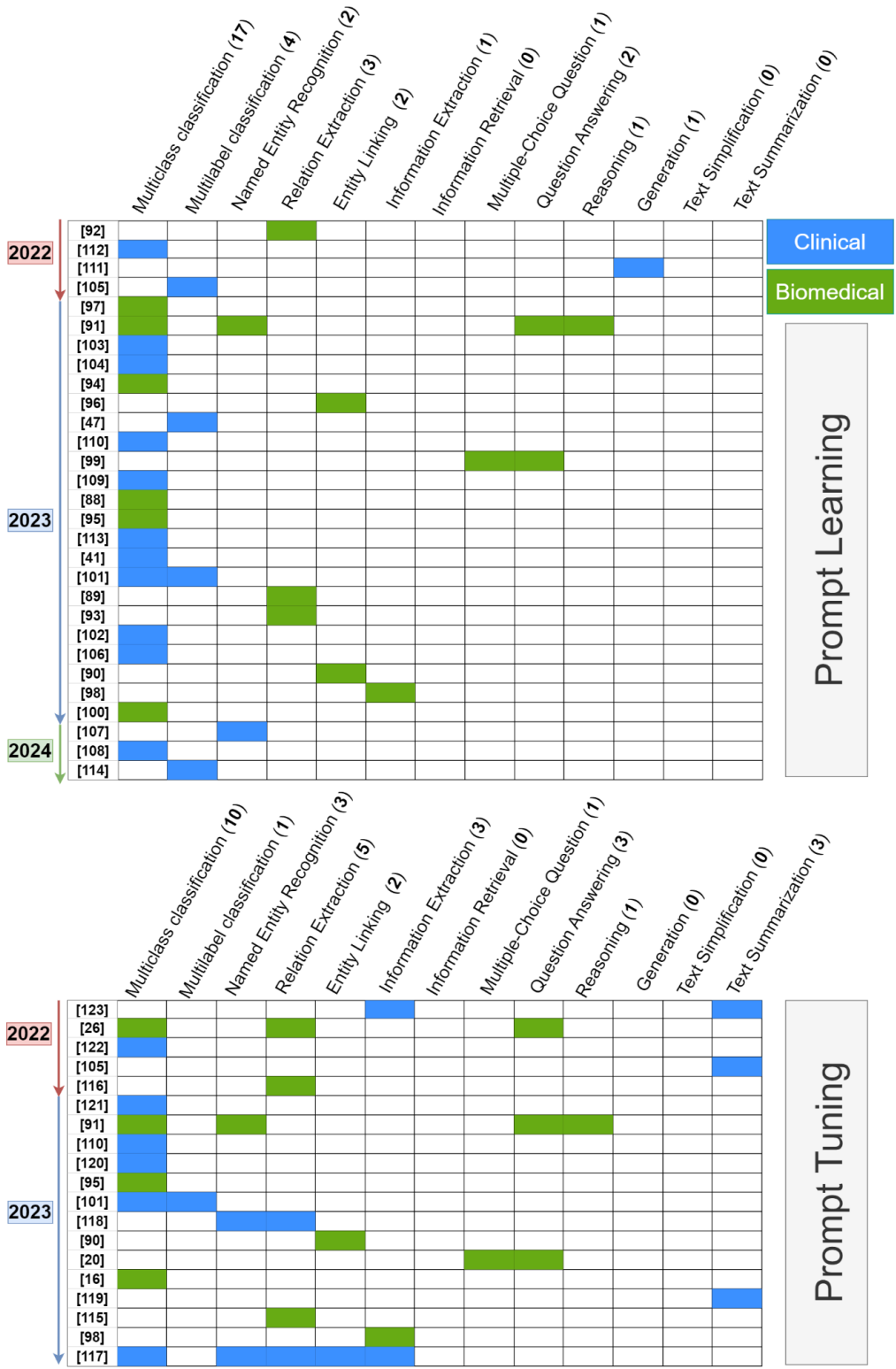

Figure 5: Overview of selected PL and PT articles, showcasing NLP tasks alongside their topic domain

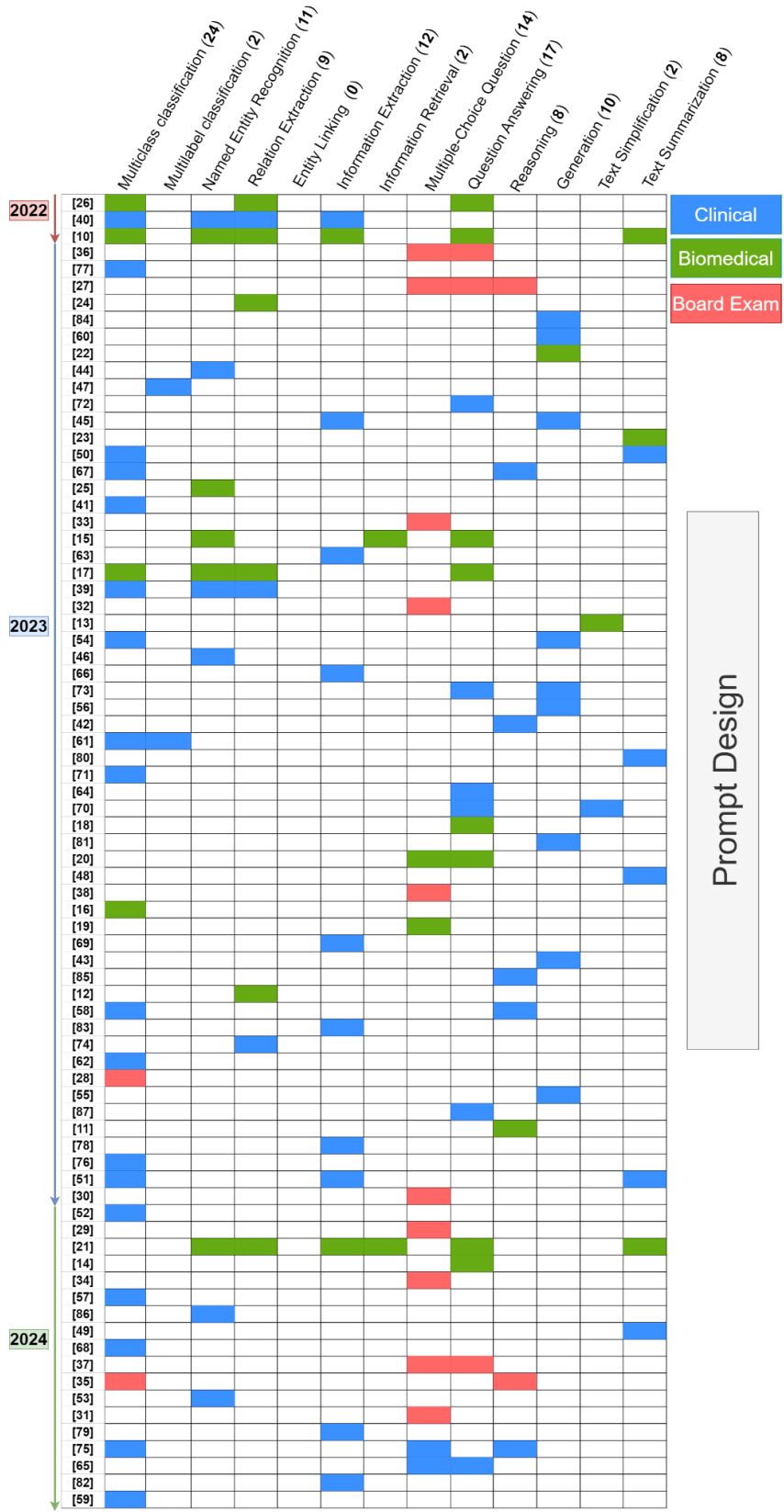

Figure 6: Overview of selected PD articles, showcasing NLP tasks alongside their topic domain

Figure 6 presents the same analysis for PD-based articles. Unlike PL and PT, a prominent trend observed is that several studies focus on real-world board exams. Notably, these studies predominantly center around tasks involving answering multiple-choice questions (MCQs). It is worth noting that although MCQs might be cast as a classification task, in practice, it is cast as a generation task using causal LLMs. It is interesting to note that none of the selected PD articles propose the task of entity linking, despite the clear opportunity of leveraging LLMs' in-context learning ability for medical entity linking.

## Prompt engineering techniques

We extensively investigated the employed prompt techniques: among PD articles, 49 studies used zero-shot prompting, 23 used few-shot prompting, and 10 used one-shot prompting. Few-shot tends to outperform in multiple-choice questions. But its advantage over zero-shot is inconsistent in other NLP tasks. We propose a comprehensive summary of the existing techniques in Table 2.

As shown in Table 2, Chain-of-Thought (CoT) prompting [2] stands as the most common technique, followed by the Persona pattern. In medical multiple-choice questions, various attempts with CoT can lead to different reasoning pathways and answers. Hence, to improve accuracy, two studies [19,20] employed self-consistency, a method involving using multiple CoT prompts and selecting the most frequently occurring answer through voting.

Flipped Interaction was used for simulation tasks, such as doctor-patient engagement [60] or to provide clinical training to medical students [81]. Emotion Enhancement was applied in mental health contexts [58,60], allowing the LLM to produce emotional statements.

More innovative prompt engineering techniques include k-nearest neighbors few-shot prompting [19] and pseudo-classification prompting [78]. The former utilizes the k-nearest neighbors algorithm to select the k closest examples in a large annotated dataset based on the input before using them in the prompt, and the latter presents to the LLMs all possible labels, asking the model to respond with a binary output for each provided label. Despite its potential, Tree-of-Thoughts pattern utilization was limited, with only one instance found among the articles [77].

Table 2: Most recurrent prompt techniques found, with the corresponding description, template and references

| Prompt techniques | Description | Prompt Template | Count articles | References |
|---|---|---|---|---|
| Chain-of-Thought (CoT) | Asking the LLM to provide the reasoning before answering. | *Think step by step.* | 17 | [11,19,20,27,29,32,33,35,39,51,58,67,75,77,82,83,85] |
| Persona (Role-defining) | Assigning the LLM a particular role to accomplish a task related to that role. | *Act as X* (e.g. *Act as a Physician, Act as a Psychiatrist,* etc). | 10 | [32,49,55,56,59–61,82,84,85] |
| Ensemble prompting | Using multiple independent prompts to answer the same question. The final output is decided by majority vote. | *Prompt1, Output1, Prompt2, Output2, [...], Promptk, Outputk Final output: Vote* | 4 | [19,20,39,52] |
| Scene-defining | Simulating a scene related to the adressed task. | *you are in a hospital, in front of a patient...* | 3 | [18,49,61] |
| Prompt-chaining | Separating a task into multiple subtasks, each resolved with a prompt. | *Prompt1->Output1, Output1+Prompt2 ->Output2, [...] Outputk-1+Promptk-> Outputk* | 3 | [37,80,84] |
| Flipped Interaction | Making the LLM take the lead (e.g. asking questions) and the user interacting with it passively. | *I would like you to ask me questions to achieve X. You should ask questions until <condition/goal> is met.* | 2 | [60,81] |
| Emotion enhancement | Making the LLM more or less expressing human-like emotions. | *You can have emotional fluctuations during the conversation.* | 2 | [58,60] |
| Prompt refinement | Using the LLM to refine the prompt such as translating the prompt, or rephrasing it. | *Please translate in English / rephrase this prompt: <P>.* | 2 | [37,48] |
| Retrieval-Augmented Generation | Combining an information retrieval component with a generative LLM. Snippets extracted from documents are fed into the system along with the input prompt to generate an enriched output. | *<List of relevant Snippets> <Input Prompt>* | 2 | [18,54] |
| Self-consistency (CoT-ensembling) | Ensemble prompting, each prompt using CoT. Ideal if a problem has many possible reasoning paths. | *CoT_Pr1, Output1, CoT_Pr2, Output2, ..., CoT_Prk, Outputk Final output: Vote* | 2 | [19,20] |

## Emerging trends

Figure 7 illustrates a chronological polar pie chart of selected articles and their citation connections, identifying five highly cited papers:

- Agrawal et al. [40] demonstrate GPT-3's clinical task performance, especially in Named Entity Recognition and Relation Extraction, through thorough prompt design.
- Kung et al. [36] evaluate ChatGPT's (GPT-3.5) ability for the US Medical Licensing Exam (USMLE), shortly after the public release of ChatGPT.
- Singhal et al. [20] introduce MultiMedQA and HealthSearchQA benchmarks. The paper also presents instruction prompt tuning for domain alignment, a novel paradigm that entails learning a soft prompt prior to the LLM general instruction, which is usually written as a hard prompt. Using this approach on FlanPaLM led to the development of Med-PaLM, improving question answering over FlanPaLM.
- Nori et al. [27] evaluate GPT-4 on the USMLE and MultiMedQA, surpassing previous state-of-the-art results, including GPT-3.5 and Med-PaLM.
- Luo et al. [26] release BioGPT, a fine-tuned variant of GPT-2 for biomedical tasks, achieving state-of-the-art results on six biomedical NLP tasks with suffix-based PT.

**Trends in Prompt design**

As shown by Figure 7, the PD paradigm presents multiple trends: all papers disseminated in clinical-based venues and 82% of the encountered preprints adhere to this paradigm. Furthermore, we discerned a notable prevalence of work on frozen LLM within the domain of PD, a trend perhaps unsurprising given the recurrent use of ChatGPT in 74 instances according to Figure 4, despite OpenAI providing fine-tuning capabilities for the model. It is worth mentioning that 64% of PD papers do not include any baseline, including human comparison. This gap will be further explored in a subsequent section.

**Trends in Prompt learning and Prompt tuning**

Among PL and PT articles, Computer Science and Medical Informatics are the most prevalent venues. Although PL has drawn attention to the idea of adapting the MLM objective to downstream tasks without needing to further update the LLM weights, many studies still opt to fine-tune their LLMs, with a non-negligible amount of them evaluating in few-shot settings [89,92,93,112]. Unlike PD, PL and PT usually include a baseline, with it often being a traditional fine-tuning version of the evaluated model [92,93,95], to compare it against novel prompt-based paradigms. These studies came to a common conclusion, being that PL is a promising alternative to traditional fine-tuning in few-shot scenarios.

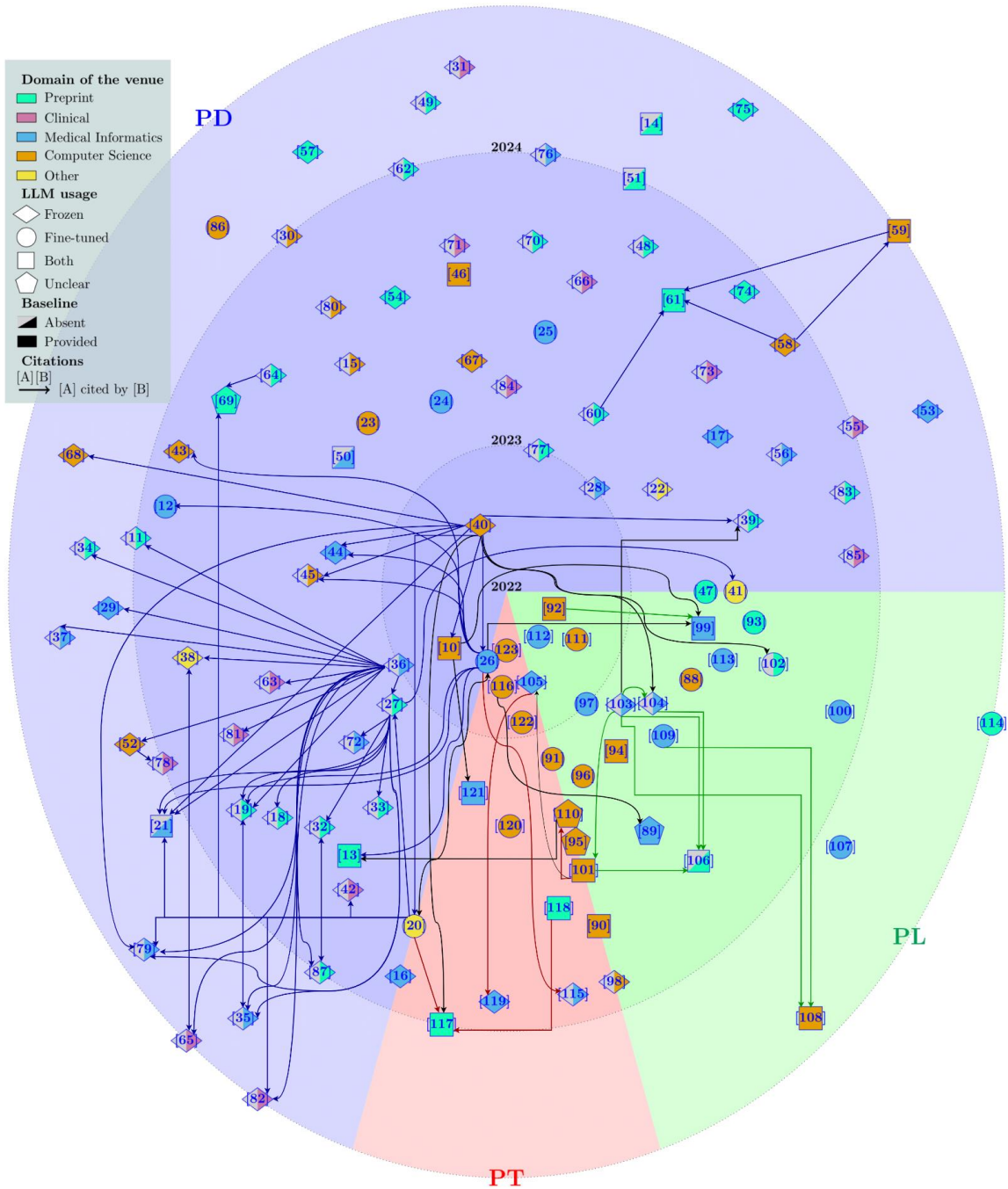

Figure 7: A chronological chart showing the selected articles, across the three prompt-based paradigms. Articles are classified by different colors according to the venues in which they were published. Different shapes illustrate whether the LLM is fine-tuned, frozen or both. Solid or striped color indicate whether authors employed a non-prompt baseline (including human) for evaluation. Arrows connecting two articles denote direct citations. The nodes in the border of PD, PL or PT are studies proposing the two involved prompt engineering paradigms.

There are two ways for conducting PL: one involves filling in the blanks within a text, known as Cloze prompts, while the other consists in predicting masked tokens at the end of the sequence, referred to as Prefix prompts. A distinct advantage of the latter approach is its compatibility with autoregressive models, as they exclusively predict the appended masks. Among the PL articles, 21 of them propose Cloze prompts, while 15 utilize Prefix prompting. The involved NLP tasks are well-distributed across these two prompt patterns. Another crucial aspect of PL is the verbalizer. As PL revolves around predicting masked tokens, classification-based tasks require mapping manually selected relevant tokens to each class (manual verbalizer). Alternatively, some studies propose a soft verbalizer, akin to soft prompts, which automatically determines the most relevant token embedding for each label through training. 16 studies explicitly mention the use of a manual verbalizer, while two explored both verbalizers to assess performance [101,110]. Only one exclusively used a soft verbalizer [89]. Another study does not use any verbalizer as it focuses on generating synthetic data by filling the blanks [111]. Notably, eight studies did not report any mention regarding the verbalizer methodology.

Regarding PT, optimal prompts are attainable through soft prompting, yet determining the appropriate soft prompt length remains obscure. Five studies tried various soft prompt lengths and reported their corresponding performances [26,105,118,119,122]. While there is no definitive optimal prompt length, a trend emerges: optimal soft prompt length typically exceeds 10 tokens. Surprisingly, eight articles omit reporting the soft prompt length. Regarding the placement of soft prompts in relation to the input and the mask, consensus is lacking. Five papers prepend the soft prompt at the input's outset, while four append it as a suffix. One article employs both strategies in a single prompt template [95]. Some innovative methods involve inserting a single soft prompt for each entity that needs to be identified in entity linking tasks, or using token-wise soft prompts, where each token in the textual input is accompanied by a distinct soft prompt. The position of soft prompts remains unreported in five studies. Finally, according to the six studies which employed mixed prompts [90,91,95,101,105,110] (a combination of hard and soft prompt), it has consistently been reported that mixed prompts lead to a better performance than hard prompts alone.

**Evaluation and comparison**

Only 62 of the screened articles reported comparisons to established baselines. These include traditional deep learning approaches (e.g. fine-tuning approach), classical machine learning algorithms (e.g. logistic regression), naive systems (e.g. majority class) or human annotation. The remaining articles solely explored prompt-related solutions, without including baseline comparisons. Figure 8 traces the presence of a non-prompt baseline among different prompt categories (8a), articles sources (8b) and NLP tasks addressed (8c).

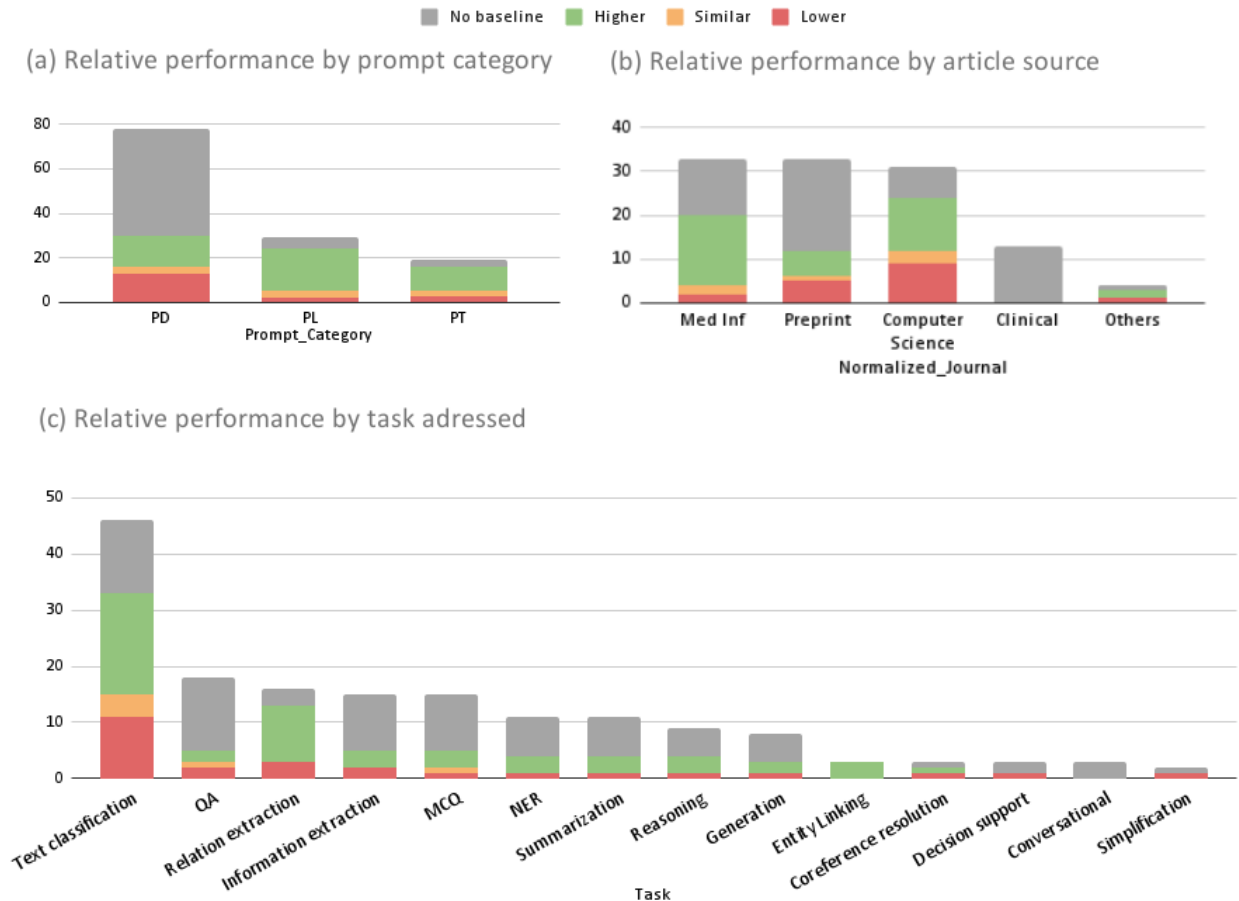

Figure 8: Baseline reports among (a) prompt categories, (b) venues, and (c) addressed NLP tasks. Higher/lower indicates that the performance of the proposed prompt-based approach is higher/lower than the baseline.

Non-prompt related baselines are often featured in studies focused on PL and PT, but not PD. Additionally, PL and PT tend to perform better than their respective reported baselines, PD tend to report less conclusive results. More specifically, among the 22 articles employing either PL or PT with an identical fine-tuned model as a baseline, 17 indicate superior performance with the prompt-based approach, 3 observed comparable performance, and 2 studies noted inferior performance.

Significantly, articles from computer science venues tend to include more state-of-the-art baselines than those from medical informatics and clinical venues. Specifically, all 13 articles reviewed from clinical venues did not utilize any non-prompt baselines. Furthermore, there appears to be no consistent link between the type of NLP tasks and the omission of baselines, indicating that the decision to include baselines is more influenced by the evaluation methodology than by feasibility.

**Prompt optimization**

Numerous studies in the literature highlight the few-shot learning capabilities of LLMs, often referred to as "few-shot prompting", wherein they demonstrate proficiency in executing tasks with minimal demonstrations provided, typically through text prompts. However, it is crucial to acknowledge that the annotation cost associated with such frameworks might extend beyond the few annotated demonstrations within the prompt. Many studies claiming to explore few-shot/zero-shot learning through prompt engineering rely on extensive annotated validation datasets to refine prompt design and formulation. This is for example the case in the article that popularized the term "few-shot learning" [1]. Among the 45 analyzed articles concentrating on few-shot/zero-shot learning, five explicitly detail the optimization of prompt formulation using extensive validation datasets. Conversely, 18 of these articles either do not engage in prompt optimization or they test various prompts and document all results. Notably, 22 articles present results using only one prompt choice, without clarifying whether this choice was made thanks to additional validation datasets.

## Discussion

**Prompt engineering techniques enable competitive performance in scenarios with limited or no resources**, as well as in environments with low-cost computing infrastructure. Figure 7 shows the absence of PL and PT-related works in clinical journals. This trend may stem from the widespread accessibility of ChatGPT, favoring PD-focused investigations. Despite efforts like OpenPrompt [125] to facilitate PL and PT works, the programming barrier likely deters clinical practitioners. Surprisingly, seven articles use ChatGPT with sensitive clinical data. Despite the recent availability of ChatGPT Enterprise in GPT-4 for secure data handling, it is apparent that most of these studies have not utilized this feature since they used GPT-3.5. Limited use of local LLMs, especially LLaMA-based, suggests a need for their increased adoption in future clinical PD studies. The lack of local LLMs may be due to clinicians' limited computational infrastructure.

In documented prompt engineering techniques, the effectiveness of few-shot prompting compared to zero-shot varies by task and scenario. However, **CoT shows superior reasoning performance**, compelling LLMs to present reasoning pathways and consistently outperforming zero-shot and few-shot methods across PD studies. Its ensemble-based variant, self-consistency, consistently outperforms CoT. Despite the Persona pattern's frequent use, there is a lack of ablation studies on its impact on medical task performance, with only one article reporting negligible improvement [61]. Prompt engineering is an emerging field of study that still needs to prove its efficacy. However, almost half of the articles focused only on prompt engineering and failed to report any non-prompt-related baseline performance, despite the availability of such baselines for the addressed NLP tasks. On the whole, the results are far from being systematically in favor of LLM-based methods, greatly attenuating the impression of a technological breakthrough that is generally commented on. Selecting a baseline remains a necessary step towards understanding the actual impact of prompt engineering.

Regarding the languages, analysis of Figure 3 underscores the dominance of English in medical literature, yet **many articles studying English fail to explicitly mention the language of study**. This oversight is more prevalent in computer science and clinical venues, whereas medical informatics exhibits a more favorable trend, as validated by a chi-squared test yielding a p-value of 0.02 (Table 4 in Supplementary Information). Notably, languages such as Chinese are consistently mentioned across the 18 selected articles. However, the Bender rule, namely *"always name the language(s) you are working on"*, seems to be well respected for languages other than English. This finding has already been documented for NLP research in general [126].

While traditional LLM fine-tuning remains a viable method for various NLP tasks, PL and PT are competitive alternatives to fine-tuning, particularly in resource-constrained and low computational scenarios. PL, leveraging pre-defined prompts to guide model behavior, offers an efficient approach in low-to-no resource environments. Conversely, PT emerges as a viable solution in low computational scenarios as it requires substantially fewer trainable parameters compared to traditional fine-tuning approaches. Since both prompt-based approaches do not require the LLM to be further trained, they are less prone to catastrophic forgetting [127].

For future research in prompt engineering, **we propose several recommendations aimed at improving research quality, reporting and reproducibility**. From this review, we identified technical and computational considerations, a lack of evaluations on baselines with a lack of ablation studies to evaluate the performance of the prompting strategies. Some studies do not clearly mention the prompt engineering choices they made. For instance, in PL, choices range from using Cloze to Prefix prompting, and from employing manual to soft verbalizer. Similarly, PT is characterized by configurations of soft prompts, such as the length, and the positions. To clarify these distinctions and enhance methodological transparency and reproducibility in future research, we have developed reporting guidelines available in Table 3. Adhering to these reporting guidelines will contribute to advancing prompt engineering methodologies and their practical applications in the medical field.

Table 3: Detailed reporting guidelines for future prompt engineering studies

| **General reporting recommendations** |
|---|
| ● The language of the study employed should be explicitly stated. |
| ● The mention of whether the LLM undergoes fine-tuning should be made explicit. |
| ● The prompt optimization process and their results should be documented to ensure transparency, whether it is through different tested manual prompts or through a validation dataset. |
| ● The terms "few-shot", "one-shot" and "zero-shot" should not be used in settings where the prompts have been optimized on annotated examples. |
| ● Experiments should include baseline comparisons or at least mention existing results, particularly when datasets originate from previous medical challenges or benchmarks. |
| **Specific to PL and PT** |
| ● Concepts (such as PL and PT) should be defined and used consistently with the consensus. |
| ● In PL experiments, the verbalizer utilized (soft, hard) should be explicitly specified or a clear justification should be provided if the verbalizer is omitted. Additionally, whether the prompt template follows the Cloze or the Prefix format should be mentioned. |
| ● In PT experiments, authors should provide details on soft prompt positions, length, and any variations tested, such as incorporating hard or mixed prompts, as part of the ablation study. |

# Methods

Our scoping review was conducted following the PRISMA-ScR guidelines for scoping reviews.

## Search strategy

We conducted a comprehensive literature search using Google Scholar, PubMed, WebOfScience, EMBase, ACL Anthology, ArXiv, MedRxiv, and BioRxiv. Four separate searches were performed using the following queries to encompass all three paradigms of prompt engineering, namely PD, PL and PT, in the medical field:

- "**Prompt Engineering**" AND (medical OR clinical OR medicine OR health OR healthcare OR biomedical)
- "**Prompt Design**" AND (medical OR clinical OR medicine OR health OR healthcare OR biomedical)
- ("**Prompt based learning**" OR "**Prompt-based learning**" OR "**Prompt learning**") AND (medical OR clinical OR medicine OR health OR healthcare OR biomedical)
- ("**Prompt tuning**" OR "**Prompt-tuning**") AND (medical OR clinical OR medicine OR health OR healthcare OR biomedical)

## Data extraction

Data extraction was conducted using *PublishOrPerish*[1] to extract Google Scholar results and *JabRef*[2] for title, abstract, and keyword searches on sources lacking this feature initially (i.e. Google Scholar and ACL). We extracted information on prompt paradigms (PD, PL, PT), involved LLMs, datasets used, studied language, domain (biomedical or clinical), medical subfield (if any), mentioned prompt engineering techniques, computational complexity, baselines, relative performances, and key findings. Additionally, we extracted journal information, and noted instances of PD/PL/PT terminology misuse. Details are available in the Supplementary Material.

## Inclusion and exclusion criteria

Studies were included if they met the following criteria:

- Focus on prompt engineering
- Involvement of at least one LLM
- Relevance to the medical field (biomedical or clinical)
- Pertaining to text-based generation (excluding vision-related prompts
- Not focusing on prompting for academic writing purposes

Furthermore, as most of the first studies about prompt engineering emerged in 2022 [2], we added the following constraint: publication date should be later than 2021.

## Screening process

The initial set of articles retrieved from the searches underwent screening based on titles, abstracts, and keywords. Screening was performed using *Rayyan.ai*[3], with two reviewers (JZ and MN) working in a double-blind process. Inter-annotator agreement was calculated, with conflicts resolved through discussion.

---

[1] https://harzing.com/resources/publish-or-perish

[2] https://www.jabref.org/

[3] https://rayyan.ai/

## Limitations

A limitation was the large number of articles retrieved during the initial search, which was addressed by limiting the search scope to titles, abstracts, and keywords. Furthermore, since some studies may perform prompt engineering techniques without mentioning any of the four prompt-related expressions employed in the queries, they might be missed by our searches.

## Acknowledgments

JZ is financed by the NCCR Evolving Language, a National Centre of Competence in Research, funded by the Swiss National Science Foundation (grant number **#51NF40_180888**).

## References

[1] T. Brown, B. Mann, N. Ryder, M. Subbiah, J.D. Kaplan, P. Dhariwal, A. Neelakantan, P. Shyam, G. Sastry, and A. Askell, Language models are few-shot learners, *Advances in Neural Information Processing Systems*. **33** (2020) 1877–1901.

[2] T. Kojima, S.S. Gu, M. Reid, Y. Matsuo, and Y. Iwasawa, Large language models are zero-shot reasoners, *Advances in Neural Information Processing Systems*. **35** (2022) 22199–22213.

[3] P. Liu, W. Yuan, J. Fu, Z. Jiang, H. Hayashi, and G. Neubig, Pre-train, Prompt, and Predict: A Systematic Survey of Prompting Methods in Natural Language Processing, *ACM Comput. Surv.* **55** (2023) 1–35. doi:10.1145/3560815.

[4] J. White, Q. Fu, S. Hays, M. Sandborn, C. Olea, H. Gilbert, A. Elnashar, J. Spencer-Smith, and D.C. Schmidt, A Prompt Pattern Catalog to Enhance Prompt Engineering with ChatGPT, *arXiv.Org*. (2023). https://arxiv.org/abs/2302.11382v1 (accessed March 14, 2024).

[5] A. Névéol, H. Dalianis, S. Velupillai, G. Savova, and P. Zweigenbaum, Clinical Natural Language Processing in languages other than English: opportunities and challenges, *J Biomed Semant*. **9** (2018) 12. doi:10.1186/s13326-018-0179-8.

[6] A.S. Luccioni, and A. Rogers, Mind your Language (Model): Fact-Checking LLMs and their Role in NLP Research and Practice, (2023). doi:10.48550/arXiv.2308.07120.

[7] A.J. Thirunavukarasu, D.S.J. Ting, K. Elangovan, L. Gutierrez, T.F. Tan, and D.S.W. Ting, Large language models in medicine, *Nat Med*. **29** (2023) 1930–1940. doi:10.1038/s41591-023-02448-8.

[8] W.X. Zhao, K. Zhou, J. Li, T. Tang, X. Wang, Y. Hou, Y. Min, B. Zhang, J. Zhang, Z. Dong, Y. Du, C. Yang, Y. Chen, Z. Chen, J. Jiang, R. Ren, Y. Li, X. Tang, Z. Liu, P. Liu, J.-Y. Nie, and J.-R. Wen, A Survey of Large Language Models, (2023). doi:10.48550/arXiv.2303.18223.

[9] B. Lester, R. Al-Rfou, and N. Constant, The Power of Scale for Parameter-Efficient Prompt Tuning, in: Proceedings of the 2021 Conference on Empirical Methods in Natural Language Processing, Association for Computational Linguistics, Online and Punta Cana, Dominican Republic, 2021: pp. 3045–3059. doi:10.18653/v1/2021.emnlp-main.243.

[10] J. Fries, L. Weber, N. Seelam, G. Altay, D. Datta, S. Garda, S. Kang, R. Su, W. Kusa, and S. Cahyawijaya, Bigbio: A framework for data-centric biomedical natural language processing, *Advances in Neural Information Processing Systems*. **35** (2022) 25792–25806.

[11] S.J. Weisenthal, ChatGPT and post-test probability, (2023). http://arxiv.org/abs/2311.12188 (accessed March 16, 2024).

[12] L. Li, and W. Ning, ProBioRE: A Framework for Biomedical Causal Relation Extraction Based on Dual-head Prompt and Prototypical Network, in: 2023 IEEE International Conference on Bioinformatics and Biomedicine (BIBM), IEEE, 2023: pp. 2071–2074. https://ieeexplore.ieee.org/abstract/document/10385919/?casa_token=1_RLqaAwXpUAAAAA:-

UydSvIHLkMDahOYoQeBSBTySh_8qAwMEBPG4CdVMi-5sF-eFnI10OtrTI6sZ61GmlQbGf-eX7X_dA (accessed March 16, 2024).

[13] Z. Li, S. Belkadi, N. Micheletti, L. Han, M. Shardlow, and G. Nenadic, Large Language Models and Control Mechanisms Improve Text Readability of Biomedical Abstracts, (2023). http://arxiv.org/abs/2309.13202 (accessed March 16, 2024).

[14] Q. Li, X. Yang, H. Wang, Q. Wang, L. Liu, J. Wang, Y. Zhang, M. Chu, S. Hu, Y. Chen, Y. Shen, C. Fan, W. Zhang, T. Xu, J. Gu, J. Zheng, and G.Z.A. Group, From Beginner to Expert: Modeling Medical Knowledge into General LLMs, (2024). http://arxiv.org/abs/2312.01040 (accessed March 16, 2024).

[15] S. Ateia, and U. Kruschwitz, Is ChatGPT a Biomedical Expert?, in: Working Notes of the Conference and Labs of the Evaluation Forum (CLEF 2023), CEUR, Thessaloniki, Greece, 2023: pp. 73–90. https://ceur-ws.org/Vol-3497/#paper-006 (accessed March 16, 2024).

[16] A. Belyaeva, J. Cosentino, F. Hormozdiari, K. Eswaran, S. Shetty, G. Corrado, A. Carroll, C.Y. McLean, and N.A. Furlotte, Multimodal LLMs for Health Grounded in Individual-Specific Data, in: Machine Learning for Multimodal Healthcare Data, Springer Nature Switzerland, Cham, 2023: pp. 86–102. doi:10.1007/978-3-031-47679-2_7.

[17] Q. Chen, H. Sun, H. Liu, Y. Jiang, T. Ran, X. Jin, X. Xiao, Z. Lin, H. Chen, and Z. Niu, An extensive benchmark study on biomedical text generation and mining with ChatGPT, *Bioinformatics*. **39** (2023) btad557.

[18] D. Mollá, Large Language Models and Prompt Engineering for Biomedical Query Focused Multi-Document Summarisation, (2023). http://arxiv.org/abs/2311.05169 (accessed March 16, 2024).

[19] H. Nori, Y.T. Lee, S. Zhang, D. Carignan, R. Edgar, N. Fusi, N. King, J. Larson, Y. Li, W. Liu, R. Luo, S.M. McKinney, R.O. Ness, H. Poon, T. Qin, N. Usuyama, C. White, and E. Horvitz, Can Generalist Foundation Models Outcompete Special-Purpose Tuning? Case Study in Medicine, (2023). doi:10.48550/arXiv.2311.16452.

[20] K. Singhal, S. Azizi, T. Tu, S.S. Mahdavi, J. Wei, H.W. Chung, N. Scales, A. Tanwani, H. Cole-Lewis, and S. Pfohl, Large language models encode clinical knowledge, *Nature*. **620** (2023) 172–180.

[21] S. Tian, Q. Jin, L. Yeganova, P.-T. Lai, Q. Zhu, X. Chen, Y. Yang, Q. Chen, W. Kim, and D.C. Comeau, Opportunities and challenges for ChatGPT and large language models in biomedicine and health, *Briefings in Bioinformatics*. **25** (2024) bbad493.

[22] S. Lim, and R. Schmälzle, Artificial intelligence for health message generation: an empirical study using a large language model (LLM) and prompt engineering, *Frontiers in Communication*. **8** (2023) 1129082.

[23] Y.-H. Wu, Y.-J. Lin, and H.-Y. Kao, IKM_Lab at BioLaySumm Task 1: Longformer-based Prompt Tuning for Biomedical Lay Summary Generation - The 22nd Workshop on Biomedical Natural Language Processing and BioNLP Shared Tasks, *The 22nd Workshop on Biomedical Natural Language Processing and BioNLP Shared Tasks*. (2023) 602–610. doi:10.18653/v1/2023.bionlp-1.64.

[24] W. Zhang, C. Chen, J. Wang, J. Liu, and T. Ruan, A co-adaptive duality-aware framework for biomedical relation extraction, *Bioinformatics*. **39** (2023). doi:10.1093/bioinformatics/btad301.

[25] P. Chen, J. Wang, H. Lin, D. Zhao, Z. Yang, and J. Wren, Few-shot biomedical named entity recognition via knowledge-guided instance generation and prompt contrastive learning, *Bioinformatics*. **39** (2023). doi:10.1093/bioinformatics/btad496.

[26] R. Luo, L. Sun, Y. Xia, T. Qin, S. Zhang, H. Poon, and T.-Y. Liu, BioGPT: generative pre-trained transformer for biomedical text generation and mining, *Briefings in Bioinformatics*. **23** (2022) bbac409. doi:10.1093/bib/bbac409.

[27] H. Nori, N. King, S.M. McKinney, D. Carignan, and E. Horvitz, Capabilities of GPT-4 on Medical Challenge Problems, (2023). http://arxiv.org/abs/2303.13375 (accessed March 16, 2024).

[28] M.V. Heinz, S. Bhattacharya, B. Trudeau, R. Quist, S.H. Song, C.M. Lee, and N.C. Jacobson, Testing domain knowledge and risk of bias of a large-scale general artificial intelligence model in mental

health, *DIGITAL HEALTH*. **9** (2023) 205520762311704. doi:10.1177/20552076231170499.

[29] Y.-T. Ting, T.-C. Hsieh, Y.-F. Wang, Y.-C. Kuo, Y.-J. Chen, P.-K. Chan, and C.-H. Kao, Performance of ChatGPT incorporated chain-of-thought method in bilingual nuclear medicine physician board examinations, *DIGITAL HEALTH*. **10** (2024) 20552076231224074. doi:10.1177/20552076231224074.

[30] S. Casola, T. Labruna, A. Lavelli, and B. Magnini, Testing ChatGPT for Stability and Reasoning: A Case Study Using Italian Medical Specialty Tests, in: Proceedings of the 9th Italian Conference on Computational Linguistics, CEUR, Venice, Italy, 2023. https://ceur-ws.org/Vol-3596/#paper13 (accessed March 16, 2024).

[31] G. Roemer, A. Li, U. Mahmood, L. Dauer, and M. Bellamy, Artificial intelligence model GPT4 narrowly fails simulated radiological protection exam, *Journal of Radiological Protection*. **44** (2024) 013502.

[32] S. Ali, O. Shahab, R. Al Shabeeb, F. Ladak, J.O. Yang, G. Nadkarni, J. Echavarria, S. Babar, A. Shaukat, and A. Soroush, General purpose large language models match human performance on gastroenterology board exam self-assessments., *medRxiv*. (2023) 2023–09.

[33] D. Patel, G. Raut, E. Zimlichman, S. Cheetirala, G. Nadkarni, B.S. Glicksberg, R. Freeman, P. Timsina, and E. Klang, The limits of prompt engineering in medical problem-solving: a comparative analysis with ChatGPT on calculation based USMLE medical questions, *medRxiv*. (2023) 2023–08.

[34] M. Sallam, K. Al-Salahat, H. Eid, J. Egger, and B. Puladi, Human versus artificial intelligence: ChatGPT-4 outperforming Bing, Bard, ChatGPT-3.5, and humans in clinical chemistry multiple-choice questions, *medRxiv*. (2024) 2024–01.

[35] T. Savage, A. Nayak, R. Gallo, E. Rangan, and J.H. Chen, Diagnostic reasoning prompts reveal the potential for large language model interpretability in medicine, *NPJ Digital Medicine*. **7** (2024) 20.

[36] T.H. Kung, M. Cheatham, A. Medenilla, C. Sillos, L. De Leon, C. Elepaño, M. Madriaga, R. Aggabao, G. Diaz-Candido, and J. Maningo, Performance of ChatGPT on USMLE: potential for AI-assisted medical education using large language models, *PLoS Digital Health*. **2** (2023) e0000198.

[37] Y. Tanaka, T. Nakata, K. Aiga, T. Etani, R. Muramatsu, S. Katagiri, H. Kawai, F. Higashino, M. Enomoto, and M. Noda, Performance of generative pretrained transformer on the national medical licensing examination in japan, *PLoS Digital Health*. **3** (2024) e0000433.

[38] M. Rosol, J.S. Gąsior, J. \Laba, K. Korzeniewski, and M. M\lyńczak, Evaluation of the performance of GPT-3.5 and GPT-4 on the Polish Medical Final Examination, *Scientific Reports*. **13** (2023) 20512.

[39] S. Sivarajkumar, M. Kelley, A. Samolyk-Mazzanti, S. Visweswaran, and Y. Wang, An Empirical Evaluation of Prompting Strategies for Large Language Models in Zero-Shot Clinical Natural Language Processing, (2023). http://arxiv.org/abs/2309.08008 (accessed March 16, 2024).

[40] M. Agrawal, S. Hegselmann, H. Lang, Y. Kim, and D. Sontag, Large language models are few-shot clinical information extractors, in: Proceedings of the 2022 Conference on Empirical Methods in Natural Language Processing, Association for Computational Linguistics, Abu Dhabi, United Arab Emirates, 2022: pp. 1998–2022. doi:10.18653/v1/2022.emnlp-main.130.

[41] B. Dong, Z. Wang, Z. Li, Z. Duan, J. Xu, T. Pan, R. Zhang, N. Liu, X. Li, and J. Wang, Toward a stable and low-resource PLM-based medical diagnostic system via prompt tuning and MoE structure, *Scientific Reports*. **13** (2023) 12595.

[42] K.L.T. Gutierrez, and P.M.L. Viacrusis, Bridging the Gap or Widening the Divide: A Call for Capacity-Building in Artificial Intelligence for Healthcare in the Philippines, *Journal of Medicine, University of Santo Tomas*. **7** (2023) 1325–1334.

[43] K.S. Islam, A.S. Nipu, P. Madiraju, and P. Deshpande, Autocompletion of Chief Complaints in the Electronic Health Records using Large Language Models, in: 2023 IEEE International Conference on Big Data (BigData), IEEE, 2023: pp. 4912–4921. https://ieeexplore.ieee.org/abstract/document/10386778/?casa_token=eV23ixpAZREAAAAA:q13zB-Zl3hAvo67B1AKKiP4oX0qlPmI-D6jKdMsFApvFhvSuILQm438u4ndi_SegSkrWvzHLsG5V3Q (accessed March 16, 2024).

[44] S. Meoni, T. Ryffel, and É. De La Clergerie, Annotate French Clinical Data Using Large Language Model Predictions, in: 2023 IEEE 11th International Conference on Healthcare Informatics (ICHI), IEEE, 2023: pp. 550–557. https://ieeexplore.ieee.org/abstract/document/10337266/?casa_token=HjaJlasCpK0AAAAA:IAul8tcswGB3V-AFDAo6rTl4xpbT08Q_3jzEoBBbMlBZfmBTcuwvel5RrBthaJ3pNX4PEb1fENFh6g (accessed March 16, 2024).

[45] S. Meoni, E. De la Clergerie, and T. Ryffel, Large language models as instructors: A study on multilingual clinical entity extraction, in: The 22nd Workshop on Biomedical Natural Language Processing and BioNLP Shared Tasks, 2023: pp. 178–190. https://aclanthology.org/2023.bionlp-1.15/ (accessed March 16, 2024).

[46] X. Wang, and Q. Yang, LingX at ROCLING 2023 MultiNER-Health Task: Intelligent Capture of Chinese Medical Named Entities by LLMs, in: Proceedings of the 35th Conference on Computational Linguistics and Speech Processing (ROCLING 2023), 2023: pp. 350–358. https://aclanthology.org/2023.rocling-1.44.pdf (accessed March 16, 2024).

[47] Y. Yang, X. Li, H. Wang, X. Li, Y. Guan, and J. Jiang, Modeling Clinical Thinking Based on Knowledge Hypergraph Attention Network and Prompt Learning for Disease Prediction, *Available at SSRN 4496800*. (2023). https://papers.ssrn.com/sol3/papers.cfm?abstract_id=4496800 (accessed March 16, 2024).

[48] Z. Yao, A. Jaafar, B. Wang, Y. Zhu, Z. Yang, and H. Yu, Do Physicians Know How to Prompt? The Need for Automatic Prompt Optimization Help in Clinical Note Generation, (2023). http://arxiv.org/abs/2311.09684 (accessed March 16, 2024).

[49] D. van Zandvoort, L. Wiersema, T. Huibers, S. van Dulmen, and S. Brinkkemper, Enhancing Summarization Performance through Transformer-Based Prompt Engineering in Automated Medical Reporting, (2024). http://arxiv.org/abs/2311.13274 (accessed March 16, 2024).

[50] B. Zhang, R. Mishra, and D. Teodoro, DS4DH at MEDIQA-Chat 2023: Leveraging SVM and GPT-3 Prompt Engineering for Medical Dialogue Classification and Summarization, in: Proceedings of the 5th Clinical Natural Language Processing Workshop, Association for Computational Linguistics, Toronto, Canada, 2023: pp. 536–545. doi:10.18653/v1/2023.clinicalnlp-1.57.

[51] W. Zhu, X. Wang, M. Chen, and B. Tang, Overview of the PromptCBLUE Shared Task in CHIP2023, (2023). http://arxiv.org/abs/2312.17522 (accessed March 16, 2024).

[52] L. Caruccio, S. Cirillo, G. Polese, G. Solimando, S. Sundaramurthy, and G. Tortora, Can ChatGPT provide intelligent diagnoses? A comparative study between predictive models and ChatGPT to define a new medical diagnostic bot, *Expert Systems with Applications*. **235** (2024) 121186.

[53] Y.-Q. Lee, C.-T. Chen, C.-C. Chen, C.-H. Lee, P. Chen, C.-S. Wu, and H.-J. Dai, Unlocking the secrets behind advanced artificial intelligence language models in deidentifying Chinese-English mixed clinical text: Development and validation study, *Journal of Medical Internet Research*. **26** (2024) e48443.

[54] R. Bhaumik, V. Srivastava, A. Jalali, S. Ghosh, and R. Chandrasekaran, Mindwatch: A smart cloud-based ai solution for suicide ideation detection leveraging large language models, *medRxiv*. (2023) 2023–09.

[55] T.F. Heston, Safety of large language models in addressing depression, *Cureus*. **15** (2023). https://www.cureus.com/articles/213293-safety-of-large-language-models-in-addressing-depression.pdf (accessed March 16, 2024).

[56] D. Grabb, The impact of prompt engineering in large language model performance: a psychiatric example, *Journal of Medical Artificial Intelligence*. **6** (2023). https://jmai.amegroups.org/article/view/8190/html (accessed March 16, 2024).

[57] W.R. dos Santos, and I. Paraboni, Prompt-based mental health screening from social media text, (2024). http://arxiv.org/abs/2401.05912 (accessed March 16, 2024).

[58] K. Yang, S. Ji, T. Zhang, Q. Xie, Z. Kuang, and S. Ananiadou, Towards Interpretable Mental Health Analysis with Large Language Models, in: Proceedings of the 2023 Conference on Empirical Methods in Natural Language Processing, Association for Computational Linguistics, Singapore, 2023: pp. 6056–6077. doi:10.18653/v1/2023.emnlp-main.370.

[59] X. Xu, B. Yao, Y. Dong, S. Gabriel, H. Yu, J. Hendler, M. Ghassemi, A.K. Dey, and D. Wang, Mental-LLM: Leveraging Large Language Models for Mental Health Prediction via Online Text Data, *Proc. ACM Interact. Mob. Wearable Ubiquitous Technol.* **8** (2024) 1–32. doi:10.1145/3643540.

[60] S. Chen, M. Wu, K.Q. Zhu, K. Lan, Z. Zhang, and L. Cui, LLM-empowered Chatbots for Psychiatrist and Patient Simulation: Application and Evaluation, (2023). http://arxiv.org/abs/2305.13614 (accessed March 16, 2024).

[61] H. Qi, Q. Zhao, J. Li, C. Song, W. Zhai, L. Dan, S. Liu, Y.J. Yu, F. Wang, and H. Zou, Supervised Learning and Large Language Model Benchmarks on Mental Health Datasets: Cognitive Distortions and Suicidal Risks in Chinese Social Media, (2023). https://www.researchsquare.com/article/rs-3523508/latest (accessed March 16, 2024).

[62] V. Sambath, Advancements of Artificial Intelligence in Mental Health Applications : A Comparative analysis of ChatGPT 3.5 and ChatGPT 4, 2023. doi:10.13140/RG.2.2.28713.36961.

[63] H.S. Choi, J.Y. Song, K.H. Shin, J.H. Chang, and B.-S. Jang, Developing prompts from large language model for extracting clinical information from pathology and ultrasound reports in breast cancer, *Radiat Oncol J*. **41** (2023) 209–216. doi:10.3857/roj.2023.00633.

[64] D.T. Lee, A. Vaid, K.M. Menon, R. Freeman, D.S. Matteson, M.P. Marin, and G.N. Nadkarni, Development of a privacy preserving large language model for automated data extraction from thyroid cancer pathology reports, (2023) 2023.11.08.23298252. doi:10.1101/2023.11.08.23298252.

[65] F. Dennstädt, J. Hastings, P.M. Putora, E. Vu, G.F. Fischer, K. Süveg, M. Glatzer, E. Riggenbach, H.-L. Hà, and N. Cihoric, Exploring Capabilities of Large Language Models such as ChatGPT in Radiation Oncology, *Advances in Radiation Oncology*. **9** (2024) 101400. doi:10.1016/j.adro.2023.101400.

[66] S. Zhu, M. Gilbert, A.I. Ghanem, F. Siddiqui, and K. Thind, Feasibility of Using Zero-Shot Learning in Transformer-Based Natural Language Processing Algorithm for Key Information Extraction from Head and Neck Tumor Board Notes, *International Journal of Radiation Oncology, Biology, Physics*. **117** (2023) e500. doi:10.1016/j.ijrobp.2023.06.1743.

[67] X. Zhao, M. Zhang, M. Ma, C. Su, Y. Liu, M. Wang, X. Qiao, J. Guo, Y. Li, and W. Ma, HW-TSC at SemEval-2023 Task 7: Exploring the Natural Language Inference Capabilities of ChatGPT and Pre-trained Language Model for Clinical Trial, in: Proceedings of the 17th International Workshop on Semantic Evaluation (SemEval-2023), Association for Computational Linguistics, Toronto, Canada, 2023: pp. 1603–1608. doi:10.18653/v1/2023.semeval-1.221.

[68] F. Nazary, Y. Deldjoo, and T. Di Noia, ChatGPT-HealthPrompt. Harnessing the Power of XAI in Prompt-Based Healthcare Decision Support using ChatGPT, in: Artificial Intelligence. ECAI 2023 International Workshops, Springer Nature Switzerland, Cham, 2024: pp. 382–397. doi:10.1007/978-3-031-50396-2_22.

[69] B. Wang, J. Lai, H. Cao, F. Jin, Q. Li, M. Tang, C. Yao, and P. Zhang, Enhancing Real-World Data Extraction in Clinical Research: Evaluating the Impact of the Implementation of Large Language Models in Hospital Setting, (2023). https://doi.org/10.21203/rs.3.rs-3644810/v2 (accessed March 16, 2024).

[70] V. Mishra, A. Sarraju, N.M. Kalwani, and J.P. Dexter, Evaluation of Prompts to Simplify Cardiovascular Disease Information Using a Large Language Model, (2023) 2023.11.08.23298225. doi:10.1101/2023.11.08.23298225.

[71] R. Feng, K.A. Brennan, Z. Azizi, J. Goyal, M. Pedron, H.J. Chang, P. Ganesan, S. Ruiperez-Campillo, B. Deb, P.L. Clopton, T. Baykaner, A.J. Rogers, and S.M. Narayan, Optimizing ChatGPT to Detect VT Recurrence From Complex Medical Notes, *Circulation*. **148** (2023) A16401–A16401.

doi:10.1161/circ.148.suppl_1.16401.
[72] M. Chowdhury, E. Lim, A. Higham, R. McKinnon, N. Ventoura, Y. He, and N. De Pennington, Can Large Language Models Safely Address Patient Questions Following Cataract Surgery?, in: Proceedings of the 5th Clinical Natural Language Processing Workshop, Association for Computational Linguistics, Toronto, Canada, 2023: pp. 131–137. doi:10.18653/v1/2023.clinicalnlp-1.17.
[73] O. Kleinig, C. Gao, J.G. Kovoor, A.K. Gupta, S. Bacchi, and W.O. Chan, How to use large language models in ophthalmology: from prompt engineering to protecting confidentiality, *Eye*. **38** (2023) 649–653. doi:10.1038/s41433-023-02772-w.
[74] V. Arsenyan, S. Bughdaryan, F. Shaya, K. Small, and D. Shahnazaryan, Large Language Models for Biomedical Knowledge Graph Construction: Information extraction from EMR notes, (2023). doi:10.48550/arXiv.2301.12473.
[75] T. Kwon, K.T. Ong, D. Kang, S. Moon, J.R. Lee, D. Hwang, Y. Sim, B. Sohn, D. Lee, and J. Yeo, Large Language Models are Clinical Reasoners: Reasoning-Aware Diagnosis Framework with Prompt-Generated Rationales, (2024). doi:10.48550/arXiv.2312.07399.
[76] C. Wang, S. Liu, A. Li, and J. Liu, Text Dialogue Analysis for Primary Screening of Mild Cognitive Impairment: Development and Validation Study, *Journal of Medical Internet Research*. **25** (2023) e51501. doi:10.2196/51501.
[77] J. Li, L. Wang, X. Chen, X. Deng, H. Wen, M. You, and W. Liu, Are You Asking GPT-4 Medical Questions Properly? - Prompt Engineering in Consistency and Reliability with Evidence-Based Guidelines for ChatGPT-4: A Pilot Stud, (2023). doi:10.21203/rs.3.rs-3336823/v1.
[78] B. Zaidat, Y.S. Lahoti, A. Yu, K.S. Mohamed, S.K. Cho, and J.S. Kim, Artificially Intelligent Billing in Spine Surgery: An Analysis of a Large Language Model, *Global Spine Journal*. (2023) 21925682231224753. doi:10.1177/21925682231224753.
[79] S. Datta, K. Lee, H. Paek, F.J. Manion, N. Ofoegbu, J. Du, Y. Li, L.-C. Huang, J. Wang, B. Lin, H. Xu, and X. Wang, AutoCriteria: a generalizable clinical trial eligibility criteria extraction system powered by large language models, *Journal of the American Medical Informatics Association*. **31** (2024) 375–385. doi:10.1093/jamia/ocad218.
[80] R. White, T. Peng, P. Sripitak, A. Rosenberg Johansen, and M. Snyder, CliniDigest: A Case Study in Large Language Model Based Large-Scale Summarization of Clinical Trial Descriptions, in: Proceedings of the 2023 ACM Conference on Information Technology for Social Good, Association for Computing Machinery, New York, NY, USA, 2023: pp. 396–402. doi:10.1145/3582515.3609559.
[81] R. Scherr, F.F. Halaseh, A. Spina, S. Andalib, and R. Rivera, ChatGPT Interactive Medical Simulations for Early Clinical Education: Case Study, *JMIR Medical Education*. **9** (2023) e49877. doi:10.2196/49877.
[82] T. Akinci D'Antonoli, A. Stanzione, C. Bluethgen, F. Vernuccio, L. Ugga, M.E. Klontzas, R. Cuocolo, R. Cannella, and B. Koçak, Large language models in radiology: fundamentals, applications, ethical considerations, risks, and future directions, *Diagn Interv Radiol*. **30** (2024) 80–90. doi:10.4274/dir.2023.232417.
[83] I.C. Wiest, D. Ferber, J. Zhu, M. van Treeck, S.K. Meyer, R. Juglan, Z.I. Carrero, D. Paech, J. Kleesiek, M.P. Ebert, D. Truhn, and J.N. Kather, From Text to Tables: A Local Privacy Preserving Large Language Model for Structured Information Retrieval from Medical Documents, (2023) 2023.12.07.23299648. doi:10.1101/2023.12.07.23299648.
[84] E. Hamed, A. Eid, and M. Alberry, Exploring ChatGPT's Potential in Facilitating Adaptation of Clinical Guidelines: A Case Study of Diabetic Ketoacidosis Guidelines, *Cureus*. **15** (2023). doi:10.7759/cureus.38784.
[85] T. Leypold, B. Schäfer, A. Boos, and J.P. Beier, Can AI Think Like a Plastic Surgeon? Evaluating GPT-4's Clinical Judgment in Reconstructive Procedures of the Upper Extremity, *Plastic and Reconstructive Surgery – Global Open*. **11** (2023) e5471. doi:10.1097/GOX.0000000000005471.

[86] J. Yang, C. Liu, W. Deng, D. Wu, C. Weng, Y. Zhou, and K. Wang, Enhancing phenotype recognition in clinical notes using large language models: PhenoBCBERT and PhenoGPT, *Patterns*. **5** (2024) 100887. doi:10.1016/j.patter.2023.100887.

[87] L. Xiong, Q. Zeng, W. Deng, W. Luo, and R. Liu, A Novel Approach to Nursing Clinical Intelligent Decision-Making: Integration of Large Language Models and Local Knowledge Base, (2023). doi:10.21203/rs.3.rs-3756467/v1.

[88] Q. Wang, C. Zeng, Y. Liu, and P. He, A Medical Question Classification Approach Based on Prompt Tuning and Contrastive Learning (S), in: International Conference on Software Engineering & Knowledge Engineering, 2023: pp. 632–635. doi:10.18293/SEKE2023-025.

[89] D. Zhao, Y. Yang, P. Chen, J. Meng, S. Sun, J. Wang, and H. Lin, Biomedical document relation extraction with prompt learning and KNN, *Journal of Biomedical Informatics*. **145** (2023) 104459. doi:10.1016/j.jbi.2023.104459.

[90] T. Zhu, Y. Qin, M. Feng, Q. Chen, B. Hu, and Y. Xiang, BioPRO: Context-Infused Prompt Learning for Biomedical Entity Linking, *IEEE/ACM Transactions on Audio, Speech, and Language Processing*. **32** (2023) 374–385. doi:10.1109/TASLP.2023.3331149.

[91] C. Liu, S. Zhang, C. Li, and H. Zhao, CPK-Adapter: Infusing Medical Knowledge into K-Adapter with Continuous Prompt, in: 2023 8th International Conference on Intelligent Computing and Signal Processing (ICSP), 2023: pp. 1017–1023. doi:10.1109/ICSP58490.2023.10248750.

[92] H.-S. Yeh, T. Lavergne, and P. Zweigenbaum, Decorate the Examples: A Simple Method of Prompt Design for Biomedical Relation Extraction, in: Proceedings of the Thirteenth Language Resources and Evaluation Conference, European Language Resources Association, Marseille, France, 2022: pp. 3780–3787. https://aclanthology.org/2022.lrec-1.403 (accessed March 16, 2024).

[93] Z. Su, X. Yu, S. Li, and P. Chen, EPTQA: A Chinese Medical Prompt Learning Method Based On Entity Pair Type Question Answering, (2023). doi:10.2139/ssrn.4563840.

[94] H. Xu, J. Zhang, Z. Wang, S. Zhang, M. Bhalerao, Y. Liu, D. Zhu, and S. Wang, GraphPrompt: Graph-Based Prompt Templates for Biomedical Synonym Prediction, *Proceedings of the AAAI Conference on Artificial Intelligence*. **37** (2023) 10576–10584. doi:10.1609/aaai.v37i9.26256.

[95] T. Chen, A. Stefanidis, Z. Jiang, and J. Su, Improving Biomedical Claim Detection using Prompt Learning Approaches, in: 2023 IEEE 4th International Conference on Pattern Recognition and Machine Learning (PRML), 2023: pp. 369–376. doi:10.1109/PRML59573.2023.10348317.

[96] Z. Xu, Y. Chen, and B. Hu, Improving Biomedical Entity Linking with Cross-Entity Interaction, *Proceedings of the AAAI Conference on Artificial Intelligence*. **37** (2023) 13869–13877. doi:10.1609/aaai.v37i11.26624.

[97] Y. Wang, Y. Wang, Z. Peng, F. Zhang, L. Zhou, and F. Yang, Medical text classification based on the discriminative pre-training model and prompt-tuning, *DIGITAL HEALTH*. **9** (2023) 20552076231193213. doi:10.1177/20552076231193213.

[98] X. Tian, P. Wang, and S. Mao, Open-World Biomedical Knowledge Probing and Verification, in: N Proceedings of The 12th International Joint Conference on Knowledge Graphs (IJCKG'23), Association for Computing Machinery, New York, NY, USA, 2023. https://ijckg2023.knowledge-graph.jp/pages/proc/paper_3.pdf.

[99] K. Lu, P. Potash, X. Lin, Y. Sun, Z. Qian, Z. Yuan, T. Naumann, T. Cai, and J. Lu, Prompt Discriminative Language Models for Domain Adaptation, in: Proceedings of the 5th Clinical Natural Language Processing Workshop, Association for Computational Linguistics, Toronto, Canada, 2023: pp. 247–258. doi:10.18653/v1/2023.clinicalnlp-1.30.

[100] Y. Hu, Y. Chen, and H. Xu, Towards More Generalizable and Accurate Sentence Classification in Medical Abstracts with Less Data, *J Healthc Inform Res*. **7** (2023) 542–556. doi:10.1007/s41666-023-00141-6.

[101] N. Taylor, Y. Zhang, D.W. Joyce, Z. Gao, A. Kormilitzin, and A. Nevado-Holgado, Clinical Prompt

Learning With Frozen Language Models, *IEEE Transactions on Neural Networks and Learning Systems*. (2023) 1–11. doi:10.1109/TNNLS.2023.3294633.

[102] I. Landi, E. Alleva, A.A. Valentine, L.A. Lepow, and A.W. Charney, Clinical Text Deduplication Practices for Efficient Pretraining and Improved Clinical Tasks, (2023). doi:10.48550/arXiv.2312.09469.

[103] S. Sivarajkumar, and Y. Wang, HealthPrompt: A Zero-shot Learning Paradigm for Clinical Natural Language Processing, *AMIA Annu Symp Proc*. **2022** (2023) 972–981.

[104] S. Sivarajkumar, and Y. Wang, Evaluation of Healthprompt for Zero-shot Clinical Text Classification, in: 2023 IEEE 11th International Conference on Healthcare Informatics (ICHI), 2023: pp. 492–494. doi:10.1109/ICHI57859.2023.00081.

[105] L. Zhang, and J. Liu, Intent-aware Prompt Learning for Medical Question Summarization, in: 2022 IEEE International Conference on Bioinformatics and Biomedicine (BIBM), 2022: pp. 672–679. doi:10.1109/BIBM55620.2022.9995317.

[106] E. Alleva, I. Landi, L.J. Shaw, E. Böttinger, T.J. Fuchs, and I. Ensari, Keyword-optimized Template Insertion for Clinical Information Extraction via Prompt-based Learning, (2023). doi:10.48550/arXiv.2310.20089.

[107] Z. Cui, K. Yu, Z. Yuan, X. Dong, and W. Luo, Language inference-based learning for Low-Resource Chinese clinical named entity recognition using language model, *Journal of Biomedical Informatics*. **149** (2024) 104559. doi:10.1016/j.jbi.2023.104559.

[108] A. Ahmed, X. Zeng, R. Xi, M. Hou, and S.A. Shah, MED-Prompt: A novel prompt engineering framework for medicine prediction on free-text clinical notes, *Journal of King Saud University - Computer and Information Sciences*. **36** (2024) 101933. doi:10.1016/j.jksuci.2024.101933.

[109] Y. Lu, X. Liu, Z. Du, Y. Gao, and G. Wang, MedKPL: A heterogeneous knowledge enhanced prompt learning framework for transferable diagnosis, *Journal of Biomedical Informatics*. **143** (2023) 104417. doi:10.1016/j.jbi.2023.104417.

[110] Y. Cui, L. Han, and G. Nenadic, MedTem2.0: Prompt-based Temporal Classification of Treatment Events from Discharge Summaries, in: Proceedings of the 61st Annual Meeting of the Association for Computational Linguistics (Volume 4: Student Research Workshop), Association for Computational Linguistics, Toronto, Canada, 2023: pp. 160–183. doi:10.18653/v1/2023.acl-srw.27.

[111] Z. Wang, and J. Sun, PromptEHR: Conditional Electronic Healthcare Records Generation with Prompt Learning, in: Proceedings of the 2022 Conference on Empirical Methods in Natural Language Processing, Association for Computational Linguistics, Abu Dhabi, United Arab Emirates, 2022: pp. 2873–2885. doi:10.18653/v1/2022.emnlp-main.185.

[112] S. Wang, L. Tang, A. Majety, J.F. Rousseau, G. Shih, Y. Ding, and Y. Peng, Trustworthy assertion classification through prompting, *Journal of Biomedical Informatics*. **132** (2022) 104139. doi:10.1016/j.jbi.2022.104139.

[113] Z. Yao, J. Tsai, W. Liu, D.A. Levy, E. Druhl, J.I. Reisman, and H. Yu, Automated identification of eviction status from electronic health record notes, *Journal of the American Medical Informatics Association*. **30** (2023) 1429–1437. doi:10.1093/jamia/ocad081.

[114] S. Kwon, X. Wang, W. Liu, E. Druhl, M.L. Sung, J.I. Reisman, W. Li, R.D. Kerns, W. Becker, and H. Yu, ODD: A Benchmark Dataset for the NLP-based Opioid Related Aberrant Behavior Detection, (2024). doi:10.48550/arXiv.2307.02591.

[115] J. Su, J. Zhang, P. Peng, and H. Wang, EGDE: A Framework for Bridging the Gap in Medical Zero-shot Relation Triplet Extraction, in: 2023 IEEE International Conference on Bioinformatics and Biomedicine (BIBM), 2023: pp. 4418–4425. doi:10.1109/BIBM58861.2023.10385666.

[116] Q. Li, Y. Wang, T. You, and Y. Lu, BioKnowPrompt: Incorporating imprecise knowledge into prompt-tuning verbalizer with biomedical text for relation extraction, *Information Sciences*. **617** (2022) 346–358. doi:10.1016/j.ins.2022.10.063.

[117]   C. Peng, X. Yang, A. Chen, Z. Yu, K.E. Smith, A.B. Costa, M.G. Flores, J. Bian, and Y. Wu, Generative Large Language Models Are All-purpose Text Analytics Engines: Text-to-text Learning Is All Your Need, (2023). doi:10.48550/arXiv.2312.06099.

[118]   C. Peng, X. Yang, K.E. Smith, Z. Yu, A. Chen, J. Bian, and Y. Wu, Model Tuning or Prompt Tuning? A Study of Large Language Models for Clinical Concept and Relation Extraction, (2023). doi:10.48550/arXiv.2310.06239.

[119]   J. Duan, F. Lu, and J. Liu, MVP: Optimizing Multi-view Prompts for Medical Dialogue Summarization, in: 2023 IEEE International Conference on Bioinformatics and Biomedicine (BIBM), 2023: pp. 500–507. doi:10.1109/BIBM58861.2023.10385916.

[120]   O. Rohanian, H. Jauncey, M. Nouriborji, V. Kumar, B.P. Gonalves, C. Kartsonaki, I. Clinical Characterisation Group, L. Merson, and D. Clifton, Using Bottleneck Adapters to Identify Cancer in Clinical Notes under Low-Resource Constraints, in: The 22nd Workshop on Biomedical Natural Language Processing and BioNLP Shared Tasks, Association for Computational Linguistics, Toronto, Canada, 2023: pp. 62–78. doi:10.18653/v1/2023.bionlp-1.5.

[121]   A. Elfrink, I. Vagliano, A. Abu-Hanna, and I. Calixto, Soft-Prompt Tuning to Predict Lung Cancer Using Primary Care Free-Text Dutch Medical Notes, in: Artificial Intelligence in Medicine, Springer Nature Switzerland, Cham, 2023: pp. 193–198. doi:10.1007/978-3-031-34344-5_23.

[122]   B.P. Singh Rawat, and H. Yu, Parameter Efficient Transfer Learning for Suicide Attempt and Ideation Detection, in: Proceedings of the 13th International Workshop on Health Text Mining and Information Analysis (LOUHI), Association for Computational Linguistics, Abu Dhabi, United Arab Emirates (Hybrid), 2022: pp. 108–115. doi:10.18653/v1/2022.louhi-1.13.

[123]   S. Xu, X. Wan, S. Hu, M. Zhou, T. Xu, H. Wang, and H. Mi, COSSUM: Towards Conversation-Oriented Structured Summarization for Automatic Medical Insurance Assessment, in: Proceedings of the 28th ACM SIGKDD Conference on Knowledge Discovery and Data Mining, Association for Computing Machinery, New York, NY, USA, 2022: pp. 4248–4256. doi:10.1145/3534678.3539116.

[124]   A. Shaitarova, J. Zaghir, A. Lavelli, M. Krauthammer, and F. Rinaldi, Exploring the Latest Highlights in Medical Natural Language Processing across Multiple Languages: A Survey, *Yearb Med Inform*. **32** (2023) 230–243. doi:10.1055/s-0043-1768726.

[125]   N. Ding, S. Hu, W. Zhao, Y. Chen, Z. Liu, H.-T. Zheng, and M. Sun, OpenPrompt: An Open-source Framework for Prompt-learning, (2021). doi:10.48550/arXiv.2111.01998.

[126]   F. Ducel, K. Fort, G. Lejeune, and Y. Lepage, Do we Name the Languages we Study? The #BenderRule in LREC and ACL articles, in: Proceedings of the Thirteenth Language Resources and Evaluation Conference, European Language Resources Association, Marseille, France, 2022: pp. 564–573. https://aclanthology.org/2022.lrec-1.60 (accessed April 16, 2024).

[127]   Y. Luo, Z. Yang, F. Meng, Y. Li, J. Zhou, and Y. Zhang, An Empirical Study of Catastrophic Forgetting in Large Language Models During Continual Fine-tuning, (2024). doi:10.48550/arXiv.2308.08747.

# Supplementary Information

## Statistical analysis

Figure 3 presents the distribution of screened articles by publication venue, categorizing them based on whether the language of study, English, is explicitly stated, inferred from figures or prompts, or not mentioned. However, the statistical relationship between the type of publication venue and the clarity of how the language of study, English, is disclosed remains ambiguous. To address this, we conducted a Chi-squared test to assess the null hypothesis: "The type of journal is independent of the manner in which the language of study, English, is disclosed, whether stated explicitly, inferred, or not mentioned at all". The found p-value is 0.0237, meaning we can reject the null hypothesis. This test was performed on all results based on the distributions with the resulting p-values available in Table 4.

Table 4: Results of the chi-squared tests performed

| Null hypothesis | Involved item | P-value result |
| --- | --- | --- |
| The **type of journal** is independent of the manner in **which the language of study, English, is disclosed**, whether stated explicitly, inferred, or not mentioned at all | Figure 3 (small plot) | **0.0237 (<0.05)** |
| The **language of study** is independent of the manner in **which it is disclosed**, whether stated explicitly, inferred, or not mentioned at all | Figure 3 (bigger plot) | **<0.0001 (<0.05)** |
| The **prompt paradigm** in the article is independent of the reported values of the **baseline** | Figure 8a | **<0.0001 (<0.05)** |
| The **type of journal** is independent of the reported values of the **baseline** | Figure 8b | **0.0006 (<0.05)** |
| The **NLP task** is independent of the reported values of the **baseline** | Figure 8c | 0.3779 (>0.05) |

## Supplementary material

Further details on the models used are available on the following link:

https://docs.google.com/spreadsheets/d/1P-d3UIRUJr_WSh01a0lNMMoMk8lu5ihb/edit?usp=sharing&ouid=103337013807472033316&rtpof=true&sd=true